\documentclass[twoside,11pt]{article}

\usepackage{obs_study_style}
\usepackage{caption}
\usepackage{subcaption}
\usepackage[table]{xcolor}
 \usepackage{ctable} 
 \usepackage{multirow}

\heading{}{}{}{}{}{Jelena Bradic and Yinchu Zhu}

\ShortHeadings{Breiman’s Two Cultures}{Bradic and Zhu}
\firstpageno{1}

\begin{document}

\title{Comments on Leo Breiman’s paper ``Statistical Modeling: The Two Cultures'' (Statistical Science, 2001, 16(3), 199-231)}

\author{\name Jelena Bradic \email jbradic@ucsd.edu \\
      \addr Department of Mathematics       and  
 Halicioglu Data Science Institute\\
     University of California, San Diego\\
 La Jolla, CA 92037, USA 
      \AND
       \name Yinchu Zhu
        \email  yinchuzhu@brandeis.edu \\
       \addr Department of Economics\\
       Brandeis University \\
       Waltham, MA 02453, USA
       }

\maketitle

\begin{abstract}Breiman challenged statisticians to think more broadly, to step into the unknown, model-free learning world, with him paving the way forward. Statistics community responded with slight optimism, some skepticism, and plenty of disbelief. Today, we are at the same crossroad anew. Faced with the enormous practical success of model-free, deep, and machine learning, we are naturally inclined to think that everything is resolved. A new frontier has emerged; the one where the role, impact, or stability of the {\it learning} algorithms is no longer measured by prediction quality, but an inferential one -- asking the questions of {\it why} and {\it if} can no longer be safely ignored.
\end{abstract}


\section{Breiman was right}

In his article ``Statistical Modeling: The Two Cultures" \citep{breiman2001statistical}, Leo Breiman marveled at the possibility of empirically built models, trained  solely to improve predictions. He argued for their potential impact on empirical applications. He advocated for a complete reversal of model-driven statistical work, the one that clumsily tries, often strenuously, to find the best or most appropriate model for a particular problem. Leo firmly believed that a new age was upon us. Age of models without models or that of algorithms tuned all so perfectly for a unique, individual, peculiar problem at hand. Looking back at it from today's perspective, with deep learning dominating the success of algorithmically-driven science, we may wonder, how is it possible that the rest of the community failed to see it? Leo was a singular voice at the time; the rest surely and steadily continued the well-established statistical modeling path.

 Breiman was a provocateur in the best possible terms. Without people like him, statistics would not be where it is today. One might argue that breakthroughs made, paradigms uncovered, premisses broken, only become possible when the well-established routes, like trenches, hard to remove, are challenged, deemed inappropriate, or invalid.

Models are well understood to be a poor, overly simplistic representation of nature, its complexity, and flexibility. However, models were believed to help approximate certain tasks useful for nature; the quote of George E.P. Box, ``all models are wrong but some are useful," is cited to this day. 
The illusion of the success of model fitting was broken with a sequence of Peter Bickel's seminar works, among others; e.g., \cite{albers1976asymptotic, bickel1993efficient}. \cite{bickel2006tailor} established that goodness of fit tests have extremely low power unless the direction of the alternative is precisely specified. Sometimes the direction of the alternative was given by the metric implicitly used when constructing the tests. Such is the case of the one-sample Kolmogorov test for goodness of fit to the uniform $(0, 1)$ distribution. It is well known that it has power at a rate $n^{-1/2}$, notably only against alternatives where $|P (X \leq 1/2)-1/2|$ is large. In the above, $n$ denotes the sample size. The $\chi^2$ tests with an increasing number of cells as $n \to \infty$, on the other hand, have trivial power in every direction at a $n^{-1/2}$ rate. 
As goodness of fit tests measure the usefulness of the developed models, these results implied impossibility in keeping up with the belief that all models are useful. 

 A new measure of success of the {\it fit }was needed, and Leo Breiman, in ``Statistical Modeling: The Two Cultures," argued, in a manner of speaking, for a new definition of a  measure of goodness of fit -- the one of predictive accuracy. A model that can predict well on a hold-out dataset was regarded as beneficial. In a way, the success of deep learning and the advent of over-parametrized neural networks are based precisely on this predictive accuracy that Leo advocated. In some communities, this measure of ``usefulness'' has nowadays overpowered all others. At the time of Breiman's article, neural networks, although present as models, were not tested and used for predictive accuracy. We understand nowadays that it is only over-parametrized, overly-complex neural network designs, with many more parameters than samples, that are regarded as most powerful predictors; see \cite{belkin2019reconciling}. Twenty years ago, Leo simply advocated, in no weak terms, that Statistics needs to take this new challenge, step-up, and do more.

\section{Statistics has since done more}

Over the last two decades, statistics has stepped outside the model-driven molds and has since Leo's call-outs done a lot more. 
Statistics has focused on achieving generalization, not through the construction of complicated models or theories, but simplification and model reduction. Among others, seminar works of \citep{tibshirani1996regression} on Lasso and \citep{fan2001variable,fan2008sure} on model selection, sparked a whole new area of interest.
Since the beginning of the 21-st century, statistics has almost entirely moved away from the goodness of fit. Instead,  prediction accuracy came into the front view. It has made strides in understanding theoretical aspects of random forests, 
 \citep{scornet2015consistency,wager2015adaptive,athey2019generalized} and boosting \citep{buhlmann2003boosting,zhang2005boosting}, in understanding prediction accuracy from scratch by establishing 
non-asymptotic prediction error bounds \citep{bickel2009simultaneous,wainwright2019high} and proposing new model-free procedures  \citep{politis2013model,barber2015controlling} as well as Bayesian random forest equivalents, such is 
Bayesian adaptive random trees (BART) \citep{chipman2010bart}, among others. 
Intersections between algorithmic-driven learning and statistics can also be seen in new pathways in building the semi-supervised inferential tools by, for example,  \cite{lafferty2007statistical} as well as by the recently departed Larry Brown and co-authors in \cite{zhang2019semi,azriel2016semi}, or others, such as  \cite{zhang2019high} and \cite{cannings2020local}. Recent interests have led to building inferential (statistical) tools around algorithm-driven learning, see, e.g., works on conformal inference by \cite{wasserman2020universal,lei2018distribution,barber2021predictive} for example. One can only hope that these are simple beginnings of a new statistical science era, driven and inspired by the interplay of model-free and model-driven learning methods.
		
	\section{Model-free and model-driven  statistics:  scrutinized and impugned}

	The practical success of machine and deep learning is perhaps the culmination of what Breiman advocated for in  the article in question,   \cite{breiman2001statistical}. Here, prediction accuracy is the sole driver of quality of success. This is, of course, exacerbated by a concurrence of both the availability of immense computing power as well by the access to datasets of previously unimaginable scale. While machine learning innovations were largely driven by Leo's proposed prediction on a hold-out data, nowadays named generalization error, the deep learning's steep success came only after a hold-out prediction fell back into a within-sample prediction.  
 At least two paradoxes emerge when contrasting Leo's work on random forests and the origins of deep learning success. 
 
 First, there is a sharp contrast between Leo's work on random forests and deep learning methods in ways they re-use the data.
 Breiman's work on random forests and the invention of bootstrap aggregation (bagging) was designed to avoid the pitfalls of re-using the same dataset. Bootstrapping the original data, combining trees drawn on different subsamples of the data, was a way of capturing different aspects of the same dataset -- all with the intention of not reusing the same data instances repeatedly. Yet, neural networks and stochastic gradient descent (SGD) training do quite the opposite. Epochs are more related to permutations than to subsampling and bagging. With SGD, same instances of the dataset are needed, required, and re-used, usually over a hundred times.
 
 Secondly, Leo was a firm advocate of model-free learning, but one could argue that neural networks are themselves quite the opposite, examples of extreme-model learning. Deep learning success can now theoretically also be prescribed to over-parametrized learning regimes, where the number of nodes and edges is required to be far greater than the number of samples; the number of layers seems secondary to the sheer number of the weight parameters \citep{ma2018power,belkin2020two}. With this view in mind,  one has to ask whether neural networks are indeed model-free methods  \citep{belkin2018understand}? In support of this hypothesis are perhaps the many papers illustrating close connections between neural networks with an infinite number of parameters and specialized,  complex  kernel ridge regression methods \citep{lee2019wide,sohl2020infinite}.

It is clear, though, that neural-network learning is not model-driven learning and that over-parametrized neural networks achieve more than models do. However, it is unclear to what purpose. 
Prediction is no longer satisfactory and is not the only measure of success in practice. Stability, reproducibility, and inference, require more than mere control of the prediction error \citep{yu2013stability,yu2020veridical}. For example, inferential tasks related to the average treatment effect are compelling and theoretically amenable to the prediction accuracy lens. Nevertheless, the impact of machine-learning methods on such tasks has yet to be unlocked.

\begin{figure}
 \centering
     \begin{subfigure}{0.49\textwidth}
     \centering
\includegraphics[width=\textwidth]{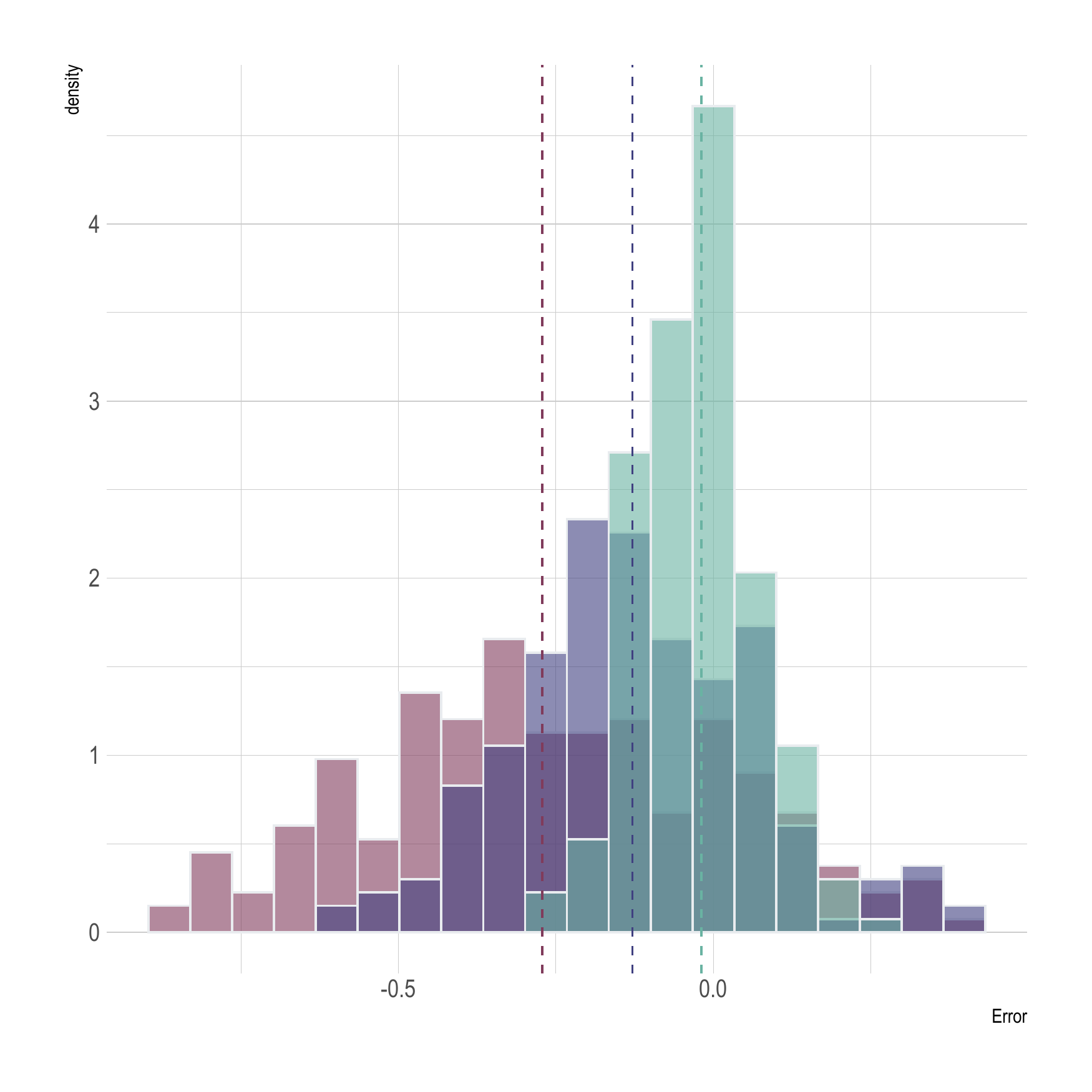}
 \caption{Covariate dimension $2$ }
         \label{fig:1}
     \end{subfigure}
     \begin{subfigure}{0.49\textwidth}
         \centering
\includegraphics[width=\textwidth]{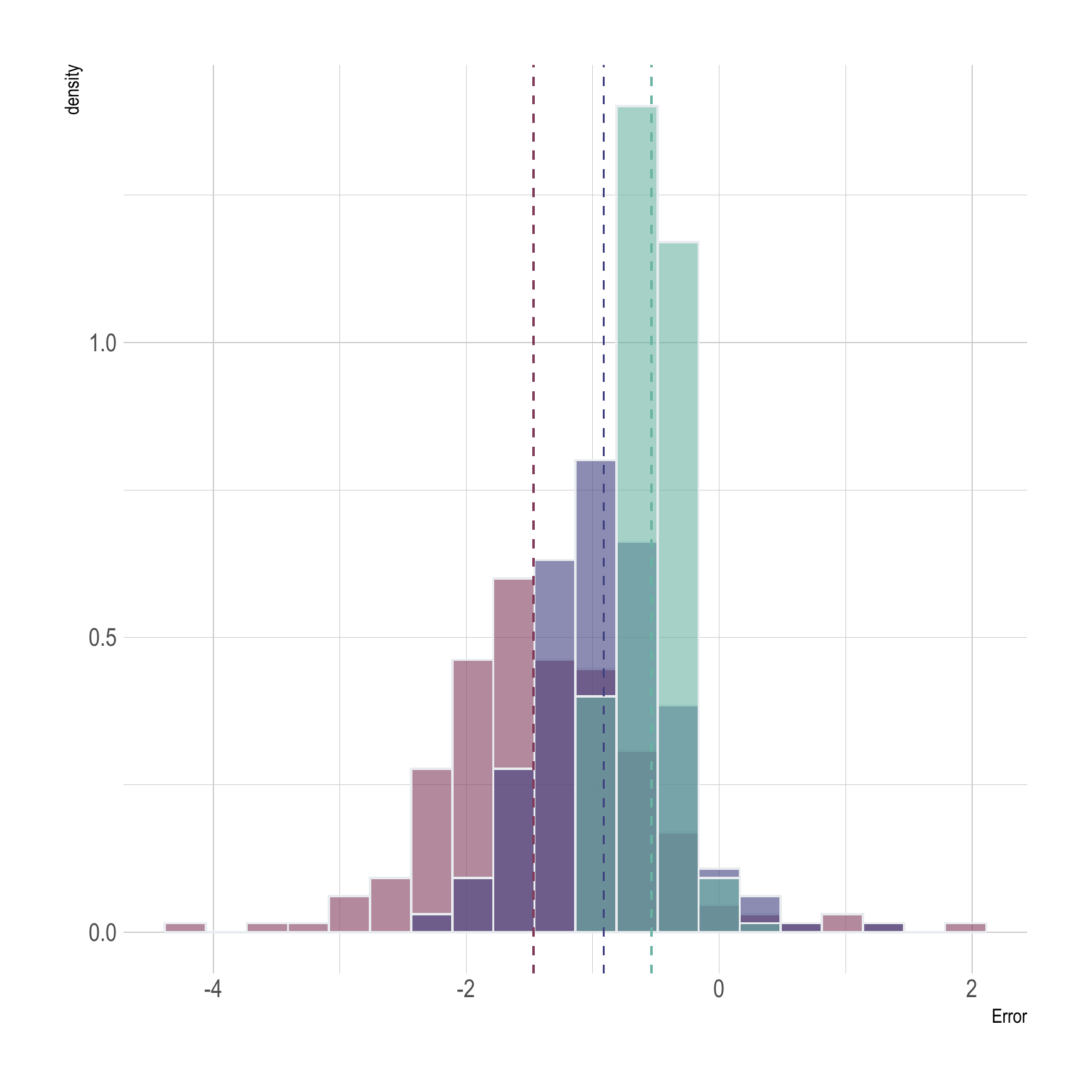}
 \caption{Covariate dimension $20$}
\label{fig:2}
     \end{subfigure}
     \caption{Histogram estimation error of $200$ repeated cross-fitted  Doubly Robust Average Treatment Effect estimator with twenty covariates across three sample sizes: $n=1000$ (pink), $n=2000$ (blue) and $n=6000$ (green). Dashed lines represent the corresponding medians. Outcome and propensity models follow Example 1.} \label{figure:1}
\end{figure}

We discuss the average treatment effect estimation and the (unknown) impact of random forest estimates for illustrative purposes. 
For brevity, we focus on a simple(st) setting of independent and identically distributed observations, receiving a binary treatment under the assumption that the treatment satisfies some form of exogeneity \citep{rosenbaum1983central}. Different forms of this assumption are referred to as unconfoundedness.
 Consider a setting with $n$ units, for whom we observe outcomes $\{Y_i\}_{i=1}^n \in \mathbb{R}$. There is a binary treatment that varies by units, denotes with $D_i \in \{0,1\}$ and a pair of potential outcomes $Y_i(0)$ and $Y_i(1)$ for all units. In this way, the realized or observed outcome can be represented as $Y_i = D_i Y_i(1) + (1-D_i) Y_i (0)$.
For each unit, we also observe a vector of potential confounders $X_i \in \mathbb R^p$.
We are interested in estimating the average treatment effect 
$\tau = \mathbb{E} [Y_i(1)-Y_i(0)],
$ which becomes identifiable under an additional overlap assumption $\mathbb{P}(D=1|X) \in (0,1)$.
 As we notice, for each unit $i$ we only observe either $Y_i(0)$ or $Y_i(1)$, making the estimation of $\tau$ challenging. There are many possible ways to estimate $\tau$, but a double robust  or augmented inverse probability weighting (AIPW) estimate stands out \citep{robins1994estimation}. It is semi-parametrically optimal \citep{hahn1998role} and asymptotically normal when either  the outcome model  or the treatment assignment model is correctly specified \citep{robins1995semiparametric} -- property also named model double-robust. There is a broad consensus that double-robustness is well understood in under-parametrized settings \citep{babino2019multiple}. With the work of \cite{10.1111/ectj.12097} double-robustness was extended to over-parametrized models, but the effect of machine-learning methods, although speculated and partially articulated, was not fully described. General conditions such as ``product-rate-condition" implied a possible validity of machine learning methods in the context of double-robust estimates. 
 
However, upon further inspection, we observe that current theoretical understandings of random forests do not extend to the ``product-rate-condition." The slow rate of convergence of the forests might indicate that a random forest estimate of the outcome and the treatment assignment might fail in satisfying this condition. 
 We performed two simple simulation experiments to explore the impact of sample size, parameter size, and the random forest itself on the ATE estimation.
Example 1 corresponds to Heterogeneous confounding with features drawn from a mixture of multivariate normal distributions with covariances being identity and Toeplitz with correlation off-diagonal being $\rho = -0.5$ and mixing probability $0.7$. Example 2 corresponds to a simple, high-dimensional, and sparse setting. Both examples are described below.
 \begin{itemize}
\item[] [Example 1.] Heterogeneous confounding
 \begin{itemize}
 \item[]Outcome: $\mathbb{E}[Y_i(a)| X_i] = X_i^\top \beta  + \log (|X_i^\top \delta|)+a, $
 
   \hskip 52pt where  $\beta  = (1,1,\dots, 1)^\top$ and $\delta  = 2 \beta$ .
 \item[]Propensity: $\mbox{logit}\Bigl[ \mathbb{P}(D_i=1|X_i)\Bigl]=  \alpha   X_i^\top \theta_1 + (1-\alpha) X_i^\top \theta_2 $ 
 
 \hskip 60pt where 
 $\theta_1= (1,1,   \dots, 1)^\top$,  $\theta_2= \theta_1/2$
  and  
 $\alpha =0.8$.
 \end{itemize}
\item[]  [Example 2.] High-dimensional and sparse confounding  
 \begin{itemize}
 \item[]Outcome: $\mathbb{E}[Y_i(a)| X_i] = X_i^\top \beta_a , $
 
  \hskip 52pt where 
  $ \beta_0 = (1,0,1,0,\dots, 0)^\top$, and $\beta_1 =(1,0,0,1,0,\dots, 0)^\top$.
 \item[]Propensity: $\mbox{logit}\Bigl[ \mathbb{P}(D_i=1|X_i)\Bigl]= X_i^\top \gamma, $
 
  \hskip 60pt where 
$\gamma= (1,1, 0, \dots, 0)^\top$.
 \end{itemize}
\end{itemize}

In the above $\mbox{logit}(q)=q/(1-q)$ and the dimensions of the features in both Examples varies from $2$, $20$ to $200$, respectively.  In Example 1 treatment assignment follows a mixture model with both models being logistic and the mixture probability being $0.8$.

\begin{figure}
 \centering
     \begin{subfigure}{0.49\textwidth}
     \centering
\includegraphics[width=\textwidth]{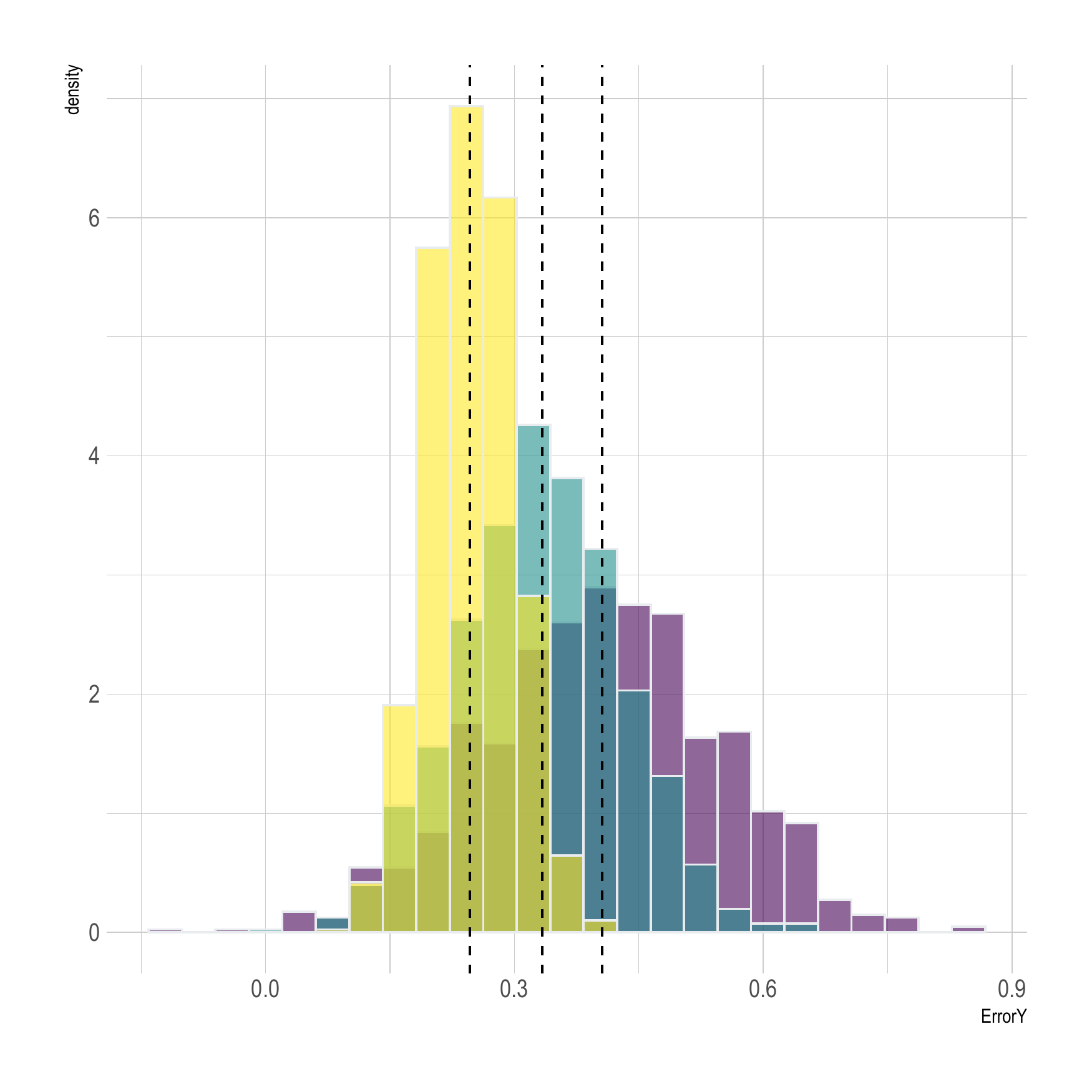}
\caption{Histogram of estimation error }
         \label{fig:3}
     \end{subfigure}
     \begin{subfigure}{0.49\textwidth}
         \centering
\includegraphics[width=\textwidth]{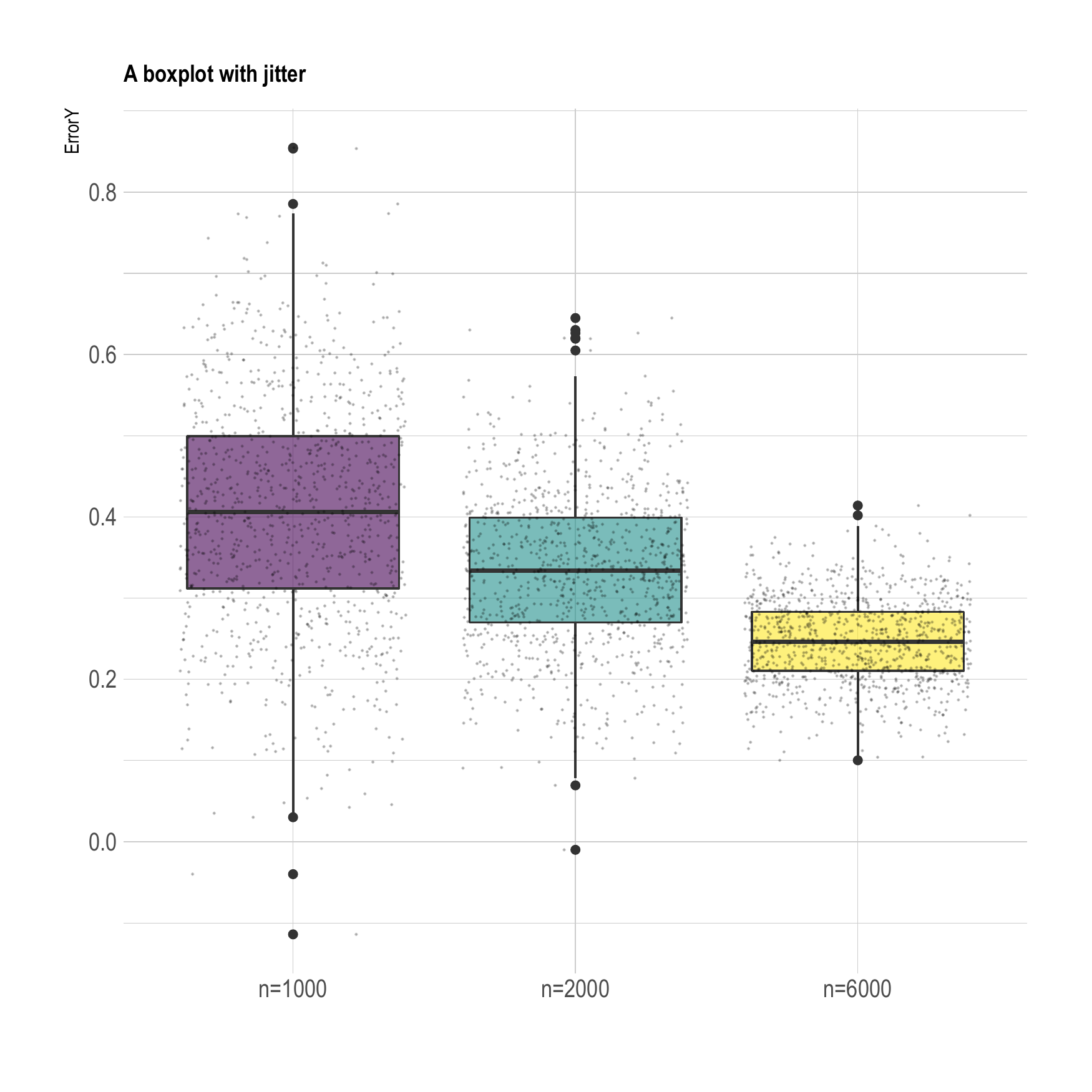}
\caption{Boxplots of estimation errors}
\label{fig:4}
     \end{subfigure}
     \caption{Estimation error of $1000$ repeated cross-fitted  Doubly Robust Average Treatment Effect estimator with $200$ covariates across three sample sizes: $n=1000$ (purple), $n=2000$ (blue) and $n=6000$ (yellow). Dashed lines represent the corresponding medians. Outcome and propensity models are $2$-sparse linear and logistic model, respectively.}\label{figure:2}
\end{figure}

We implement a random-forest version of the cross-fitted augmented inverse propensity score estimator. The outcome and propensities are estimated in a single sample, $I_1$ then evaluated on a separate sample, $I_2$, and averaged. Honest trees were used, whereas tuning parameters of the random forest were trained using cross-validation.
Let $ \widehat{ \mu_a} (\cdot)$ and $\widehat{ \pi_a}(\cdot)$ denote the estimated random forests corresponding to the potential outcome models $\mathbb{E}[Y_i(a)| X_i=x] = \mu_a(x)$ as well as the treatment assignment model $ \widehat{ \mathbb{P}}[D_i =a| X_i=x] =\pi_a(x)$.
The AIPW cross-fitted estimate is then neatly represented as 
\[
\hat \tau = \frac{1}{|I_2|} \sum_{i \in I_2} \Biggl ( \widehat{ \mu_1} (X_i)- \widehat{ \mu_0} (X_i)+ \frac{D_i - \widehat{ \pi_1}(X_i)}{\widehat{ \pi_1}(X_i)  (1-\widehat{ \pi_1}(X_i) )} \left( Y_i - \widehat{ \pi }_{D_i}(X_i)\right) \Biggl).
\]

We varied the sample size from $n=1000$ to $n=2000$ and $n=6000$. Plots highlighting the histograms and boxplots of the error $\hat \tau -1$ of the estimated average treatment effect are presented in Figures \ref{figure:1} and \ref{figure:2}. Example 1 is presented in Figures \ref{fig:1} and \ref{fig:2} while Example 2 in Figures \ref{fig:3} and \ref{fig:4}.

 The results are clear. Random forest, doubly-robust estimate fails to cover the true effect in almost all instances. 
 Average coverage corresponding to Example 1 and Example 2 are presented in Table \ref{table:1}.   In low-dimensional problems, with $p=2$ the case of large sample size $n=6000$ gets it close to covering but  even with $p=20$ the coverage is far from $95\%$.  For covariate dimension $200$, we see an awkward false concentration below zero: the estimate is getting more sure, less variable but at the wrong center. 
 
 \newcolumntype{?}{!{\vrule width 1.5pt}}
\renewcommand{\arraystretch}{1.5}



\begin{table}[h!]
\centering
\begin{tabular}{p{2cm} ? p{2cm}| p{2cm}p{2cm}p{2cm} }
 & &$n=1000$& $n=2000$ & $n=6000$ \\
\specialrule{.2em}{.2em}{0em}
 \multirow{2}{5em}{Example 1}
 & $p=2$&\textcolor{black}{83.5}  &\textcolor{black}{89.5} & \cellcolor[HTML]{69b3a2} \textcolor{black}{94}   \\
&  $p=20$& \textcolor{black}{56.6}   &\textcolor{black}{58.5}&\textcolor{black}{32.5} \\
\hline
 Example 2& $p=200$&  {\textcolor{black}{18.5}}   & {\textcolor{black}{07}} &  {\textcolor{black}{0.1}} \\
\end{tabular}
\caption{Coverage of $95\%$ confidence intervals.}
\label{table:1}
\end{table}

 The reasons for these shortcomings are unknown at the moment. We need to further our knowledge of model-free learning' effects beyond their sole predictive accuracy.

\section{The next frontier:  theory {\it with} practice instead of theory {\it vs.} practice}

The new age of Data Science is upon us. With it comes the new challenge of addressing the questions of {\it why} and {\it if} a scientific or practical phenomenon has been discovered. This, in turn, requires new standards, formulations, definitions aimed at addressing the fundamental questions of whether there exists a phenomenon to be discovered, why the discovery was made, who influenced it, will it change drastically if we were to have observed somewhat different, distorted or data that has been intervened upon. Although models are known to be a poor representation of nature, we are now faced with questions whether our current, well-established, and natural go-to definitions are an excessively simplified depiction of practice, of the type of questions that might influence the domain of applications significantly. One might wonder if we now have to think of advancing the questions rather than models so that they advantageously drive the science?

Moreover, instinctive theoretical principles are no longer valid.  
Nonparametrics, an area perhaps the closest to model-free learning, traditionally utilizes different regularization methods to stabilize the estimators. Paradoxically, Tikhonov regularization \citep{tikhonov1943stability} characterizes much of the early nonparametrics works \citep{hoerl1970ridge}, and it also characterizes current theoretical underpinnings of deep neural networks; e.g., \cite{jacot2018neural,arora2019harnessing}. More broadly, regularization has been one of the threads underlying and connecting various research areas of statistics in the past two decades. We have made strong strides in understanding it, using it, designing it. We've successfully designed inferential tasks on the shoulders  of those findings despite the negative impact of regularization, that is, the bias it affects; e.g.,\cite{van2014asymptotically,zhu2018significance}. Regularization, albeit more implicit, is ever-present in deep learning. Much of the practical success of neural networks is attributed to various effects of the regularization. Suddenly the effect of regularization is multi-faceted: parameter, architecture, batch normalization, gradient descent, and more \citep{neyshabur2017geometry,razin2020implicit}. This begs the question of whether the established notion of regularization is perhaps too wide to be useful?
 
 Leo advocated strongly in favor of model-free learning. One of the natural bridges between model-free and model-driven learning lies perhaps in the notion of model misspecification. However, we now understand that model misspecification manifests itself differently in under- and over-parametrized settings \citep{bradic2019minimax}. Classical notions reflect misspecification concerning parametrization structure, but we now understand that misspecification in the number of parameters is also possible. Fundamental limits of these impacts are largely unknown, although some progress has been made, for example, in \cite{bradic2018testability}, \cite{bradic2019sparsity} or \cite{cai2018accuracy} where robustness to sparsity is studied.  Since the effect of the classical definition changes, new definitions are needed even for model misspecification. We see that robustness guarantees depend on specific structures which we do not understand well yet. Inferential tasks inadvertently suffer because of it, and we still do not understand the reason behind it all.

Lastly, new formulations are needed to better understand the new data world surrounding us,  which uses the data to make decisions affecting millions of people, benefit science, and push the boundaries of the existing domains. Perhaps it is again the time to lean on Leo's firm conviction that breakthroughs happen in conjunction with the practical applications and not against them. It is the time to listen and interact with practice, learn how to ask better questions, not only provide a better fitting but bridge the gap between theoretical goals and their purpose, usefulness, and scientific discovery.


\acks{We would like to acknowledge support  of NSF  DMS award number  \#1712481.}


\newpage






\vskip 0.2in
\bibliography{Bradic_Zhu}

\begin{thebibliography}{51}
\providecommand{\natexlab}[1]{#1}
\providecommand{\url}[1]{\texttt{#1}}
\expandafter\ifx\csname urlstyle\endcsname\relax
  \providecommand{\doi}[1]{doi: #1}\else
  \providecommand{\doi}{doi: \begingroup \urlstyle{rm}\Url}\fi

\bibitem[Albers et~al.(1976)Albers, Bickel, and van Zwet]{albers1976asymptotic}
Willem Albers, Peter~J Bickel, and Willem~R van Zwet.
\newblock Asymptotic expansions for the power of distribution free tests in the
  one-sample problem.
\newblock \emph{The Annals of Statistics}, pages 108--156, 1976.

\bibitem[Arora et~al.(2019)Arora, Du, Li, Salakhutdinov, Wang, and
  Yu]{arora2019harnessing}
Sanjeev Arora, Simon~S Du, Zhiyuan Li, Ruslan Salakhutdinov, Ruosong Wang, and
  Dingli Yu.
\newblock Harnessing the power of infinitely wide deep nets on small-data
  tasks.
\newblock \emph{arXiv preprint arXiv:1910.01663}, 2019.

\bibitem[Athey et~al.(2019)Athey, Tibshirani, and Wager]{athey2019generalized}
Susan Athey, Julie Tibshirani, and Stefan Wager.
\newblock Generalized random forests.
\newblock \emph{Annals of Statistics}, 47\penalty0 (2):\penalty0 1148--1178,
  2019.

\bibitem[Azriel et~al.(2016)Azriel, Brown, Sklar, Berk, Buja, and
  Zhao]{azriel2016semi}
David Azriel, Lawrence~D Brown, Michael Sklar, Richard Berk, Andreas Buja, and
  Linda Zhao.
\newblock Semi-supervised linear regression.
\newblock \emph{arXiv preprint arXiv:1612.02391}, 2016.

\bibitem[Babino et~al.(2019)Babino, Rotnitzky, and Robins]{babino2019multiple}
Lucia Babino, Andrea Rotnitzky, and James Robins.
\newblock Multiple robust estimation of marginal structural mean models for
  unconstrained outcomes.
\newblock \emph{Biometrics}, 75\penalty0 (1):\penalty0 90--99, 2019.

\bibitem[Barber and Cand{\`e}s(2015)]{barber2015controlling}
Rina~Foygel Barber and Emmanuel~J Cand{\`e}s.
\newblock Controlling the false discovery rate via knockoffs.
\newblock \emph{Annals of Statistics}, 43\penalty0 (5):\penalty0 2055--2085,
  2015.

\bibitem[Barber et~al.(2021)Barber, Candes, Ramdas, and
  Tibshirani]{barber2021predictive}
Rina~Foygel Barber, Emmanuel~J Candes, Aaditya Ramdas, and Ryan~J Tibshirani.
\newblock Predictive inference with the jackknife+.
\newblock \emph{The Annals of Statistics}, 49\penalty0 (1):\penalty0 486--507,
  2021.

\bibitem[Belkin et~al.(2018)Belkin, Ma, and Mandal]{belkin2018understand}
Mikhail Belkin, Siyuan Ma, and Soumik Mandal.
\newblock To understand deep learning we need to understand kernel learning.
\newblock In \emph{International Conference on Machine Learning}, pages
  541--549. PMLR, 2018.

\bibitem[Belkin et~al.(2019)Belkin, Hsu, Ma, and Mandal]{belkin2019reconciling}
Mikhail Belkin, Daniel Hsu, Siyuan Ma, and Soumik Mandal.
\newblock Reconciling modern machine-learning practice and the classical
  bias--variance trade-off.
\newblock \emph{Proceedings of the National Academy of Sciences}, 116\penalty0
  (32):\penalty0 15849--15854, 2019.

\bibitem[Belkin et~al.(2020)Belkin, Hsu, and Xu]{belkin2020two}
Mikhail Belkin, Daniel Hsu, and Ji~Xu.
\newblock Two models of double descent for weak features.
\newblock \emph{SIAM Journal on Mathematics of Data Science}, 2\penalty0
  (4):\penalty0 1167--1180, 2020.

\bibitem[Bickel et~al.(2006)Bickel, Ritov, and Stoker]{bickel2006tailor}
Peter~J Bickel, Ya\'acov Ritov, and Thomas~M. Stoker.
\newblock Tailor-made tests for goodness of fit to semiparametric hypotheses.
\newblock \emph{The Annals of Statistics}, 34\penalty0 (2):\penalty0 721--741,
  2006.

\bibitem[Bickel et~al.(2009)Bickel, Ritov, and
  Tsybakov]{bickel2009simultaneous}
Peter~J. Bickel, Ya’acov Ritov, and Alexandre~B. Tsybakov.
\newblock Simultaneous analysis of lasso and dantzig selector.
\newblock \emph{The Annals of statistics}, 37\penalty0 (4):\penalty0
  1705--1732, 2009.

\bibitem[Bickel et~al.(1993)Bickel, Klaassen, Ritov, and
  Wellner]{bickel1993efficient}
PJ~Bickel, CAJ Klaassen, Y~Ritov, and JA~Wellner.
\newblock Efficient and adaptive inference in semiparametric models, 1993.

\bibitem[Bradic et~al.(2018)Bradic, Fan, and Zhu]{bradic2018testability}
Jelena Bradic, Jianqing Fan, and Yinchu Zhu.
\newblock Testability of high-dimensional linear models with non-sparse
  structures.
\newblock \emph{arXiv preprint arXiv:1802.09117}, 2018.

\bibitem[Bradic et~al.(2019{\natexlab{a}})Bradic, Chernozhukov, Newey, and
  Zhu]{bradic2019minimax}
Jelena Bradic, Victor Chernozhukov, Whitney~K Newey, and Yinchu Zhu.
\newblock Minimax semiparametric learning with approximate sparsity.
\newblock \emph{arXiv preprint arXiv:1912.12213}, 2019{\natexlab{a}}.

\bibitem[Bradic et~al.(2019{\natexlab{b}})Bradic, Wager, and
  Zhu]{bradic2019sparsity}
Jelena Bradic, Stefan Wager, and Yinchu Zhu.
\newblock Sparsity double robust inference of average treatment effects.
\newblock \emph{arXiv preprint arXiv:1905.00744}, 2019{\natexlab{b}}.

\bibitem[Breiman(2001)]{breiman2001statistical}
Leo Breiman.
\newblock Statistical modeling: The two cultures (with comments and a rejoinder
  by the author).
\newblock \emph{Statistical science}, 16\penalty0 (3):\penalty0 199--231, 2001.

\bibitem[B{\"u}hlmann and Yu(2003)]{buhlmann2003boosting}
Peter B{\"u}hlmann and Bin Yu.
\newblock Boosting with the l 2 loss: regression and classification.
\newblock \emph{Journal of the American Statistical Association}, 98\penalty0
  (462):\penalty0 324--339, 2003.

\bibitem[Cai and Guo(2018)]{cai2018accuracy}
T~Tony Cai and Zijian Guo.
\newblock Accuracy assessment for high-dimensional linear regression.
\newblock \emph{Annals of Statistics}, 46\penalty0 (4):\penalty0 1807--1836,
  2018.

\bibitem[Cannings et~al.(2020)Cannings, Berrett, and
  Samworth]{cannings2020local}
Timothy~I Cannings, Thomas~B Berrett, and Richard~J Samworth.
\newblock Local nearest neighbour classification with applications to
  semi-supervised learning.
\newblock \emph{Annals of Statistics}, 48\penalty0 (3):\penalty0 1789--1814,
  2020.

\bibitem[Chernozhukov et~al.(2018)Chernozhukov, Chetverikov, Demirer, Duflo,
  Hansen, Newey, and Robins]{10.1111/ectj.12097}
Victor Chernozhukov, Denis Chetverikov, Mert Demirer, Esther Duflo, Christian
  Hansen, Whitney Newey, and James Robins.
\newblock {Double/debiased machine learning for treatment and structural
  parameters}.
\newblock \emph{The Econometrics Journal}, 21\penalty0 (1):\penalty0 C1--C68,
  01 2018.

\bibitem[Chipman et~al.(2010)Chipman, George, and McCulloch]{chipman2010bart}
Hugh~A Chipman, Edward~I George, and Robert~E McCulloch.
\newblock Bart: Bayesian additive regression trees.
\newblock \emph{The Annals of Applied Statistics}, 4\penalty0 (1):\penalty0
  266--298, 2010.

\bibitem[Fan and Li(2001)]{fan2001variable}
Jianqing Fan and Runze Li.
\newblock Variable selection via nonconcave penalized likelihood and its oracle
  properties.
\newblock \emph{Journal of the American statistical Association}, 96\penalty0
  (456):\penalty0 1348--1360, 2001.

\bibitem[Fan and Lv(2008)]{fan2008sure}
Jianqing Fan and Jinchi Lv.
\newblock Sure independence screening for ultrahigh dimensional feature space.
\newblock \emph{Journal of the Royal Statistical Society: Series B (Statistical
  Methodology)}, 70\penalty0 (5):\penalty0 849--911, 2008.

\bibitem[Hahn(1998)]{hahn1998role}
Jinyong Hahn.
\newblock On the role of the propensity score in efficient semiparametric
  estimation of average treatment effects.
\newblock \emph{Econometrica}, pages 315--331, 1998.

\bibitem[Hoerl and Kennard(1970)]{hoerl1970ridge}
Arthur~E Hoerl and Robert~W Kennard.
\newblock Ridge regression: Biased estimation for nonorthogonal problems.
\newblock \emph{Technometrics}, 12\penalty0 (1):\penalty0 55--67, 1970.

\bibitem[Jacot et~al.(2018)Jacot, Gabriel, and Hongler]{jacot2018neural}
Arthur Jacot, Franck Gabriel, and Cl{\'e}ment Hongler.
\newblock Neural tangent kernel: Convergence and generalization in neural
  networks.
\newblock \emph{arXiv preprint arXiv:1806.07572}, 2018.

\bibitem[Lafferty and Wasserman(2007)]{lafferty2007statistical}
John Lafferty and Larry Wasserman.
\newblock Statistical analysis of semi-supervised regression.
\newblock 2007.

\bibitem[Lee et~al.(2019)Lee, Xiao, Schoenholz, Bahri, Novak, Sohl-Dickstein,
  and Pennington]{lee2019wide}
Jaehoon Lee, Lechao Xiao, Samuel~S Schoenholz, Yasaman Bahri, Roman Novak,
  Jascha Sohl-Dickstein, and Jeffrey Pennington.
\newblock Wide neural networks of any depth evolve as linear models under
  gradient descent.
\newblock \emph{arXiv preprint arXiv:1902.06720}, 2019.

\bibitem[Lei et~al.(2018)Lei, G’Sell, Rinaldo, Tibshirani, and
  Wasserman]{lei2018distribution}
Jing Lei, Max G’Sell, Alessandro Rinaldo, Ryan~J Tibshirani, and Larry
  Wasserman.
\newblock Distribution-free predictive inference for regression.
\newblock \emph{Journal of the American Statistical Association}, 113\penalty0
  (523):\penalty0 1094--1111, 2018.

\bibitem[Ma et~al.(2018)Ma, Bassily, and Belkin]{ma2018power}
Siyuan Ma, Raef Bassily, and Mikhail Belkin.
\newblock The power of interpolation: Understanding the effectiveness of sgd in
  modern over-parametrized learning.
\newblock In \emph{International Conference on Machine Learning}, pages
  3325--3334. PMLR, 2018.

\bibitem[Neyshabur et~al.(2017)Neyshabur, Tomioka, Salakhutdinov, and
  Srebro]{neyshabur2017geometry}
Behnam Neyshabur, Ryota Tomioka, Ruslan Salakhutdinov, and Nathan Srebro.
\newblock Geometry of optimization and implicit regularization in deep
  learning.
\newblock \emph{arXiv preprint arXiv:1705.03071}, 2017.

\bibitem[Politis(2013)]{politis2013model}
Dimitris~N Politis.
\newblock Model-free model-fitting and predictive distributions.
\newblock \emph{Test}, 22\penalty0 (2):\penalty0 183--221, 2013.

\bibitem[Razin and Cohen(2020)]{razin2020implicit}
Noam Razin and Nadav Cohen.
\newblock Implicit regularization in deep learning may not be explainable by
  norms.
\newblock \emph{arXiv preprint arXiv:2005.06398}, 2020.

\bibitem[Robins and Rotnitzky(1995)]{robins1995semiparametric}
James~M Robins and Andrea Rotnitzky.
\newblock Semiparametric efficiency in multivariate regression models with
  missing data.
\newblock \emph{Journal of the American Statistical Association}, 90\penalty0
  (429):\penalty0 122--129, 1995.

\bibitem[Robins et~al.(1994)Robins, Rotnitzky, and Zhao]{robins1994estimation}
James~M Robins, Andrea Rotnitzky, and Lue~Ping Zhao.
\newblock Estimation of regression coefficients when some regressors are not
  always observed.
\newblock \emph{Journal of the American statistical Association}, 89\penalty0
  (427):\penalty0 846--866, 1994.

\bibitem[Rosenbaum and Rubin(1983)]{rosenbaum1983central}
Paul~R Rosenbaum and Donald~B Rubin.
\newblock The central role of the propensity score in observational studies for
  causal effects.
\newblock \emph{Biometrika}, 70\penalty0 (1):\penalty0 41--55, 1983.

\bibitem[Scornet et~al.(2015)Scornet, Biau, and Vert]{scornet2015consistency}
Erwan Scornet, G{\'e}rard Biau, and Jean-Philippe Vert.
\newblock Consistency of random forests.
\newblock \emph{The Annals of Statistics}, 43\penalty0 (4):\penalty0
  1716--1741, 2015.

\bibitem[Sohl-Dickstein et~al.(2020)Sohl-Dickstein, Novak, Schoenholz, and
  Lee]{sohl2020infinite}
Jascha Sohl-Dickstein, Roman Novak, Samuel~S Schoenholz, and Jaehoon Lee.
\newblock On the infinite width limit of neural networks with a standard
  parameterization.
\newblock \emph{arXiv preprint arXiv:2001.07301}, 2020.

\bibitem[Tibshirani(1996)]{tibshirani1996regression}
Robert Tibshirani.
\newblock Regression shrinkage and selection via the lasso.
\newblock \emph{Journal of the Royal Statistical Society: Series B
  (Methodological)}, 58\penalty0 (1):\penalty0 267--288, 1996.

\bibitem[Tikhonov(1943)]{tikhonov1943stability}
Andrey~Nikolayevich Tikhonov.
\newblock On the stability of inverse problems.
\newblock In \emph{Dokl. Akad. Nauk SSSR}, volume~39, pages 195--198, 1943.

\bibitem[Van~de Geer et~al.(2014)Van~de Geer, B{\"u}hlmann, Ritov, and
  Dezeure]{van2014asymptotically}
Sara Van~de Geer, Peter B{\"u}hlmann, Ya’acov Ritov, and Ruben Dezeure.
\newblock On asymptotically optimal confidence regions and tests for
  high-dimensional models.
\newblock \emph{Annals of Statistics}, 42\penalty0 (3):\penalty0 1166--1202,
  2014.

\bibitem[Wager and Walther(2015)]{wager2015adaptive}
Stefan Wager and Guenther Walther.
\newblock Adaptive concentration of regression trees, with application to
  random forests.
\newblock \emph{arXiv preprint arXiv:1503.06388}, 2015.

\bibitem[Wainwright(2019)]{wainwright2019high}
Martin~J Wainwright.
\newblock \emph{High-dimensional statistics: A non-asymptotic viewpoint},
  volume~48.
\newblock Cambridge University Press, 2019.

\bibitem[Wasserman et~al.(2020)Wasserman, Ramdas, and
  Balakrishnan]{wasserman2020universal}
Larry Wasserman, Aaditya Ramdas, and Sivaraman Balakrishnan.
\newblock Universal inference.
\newblock \emph{Proceedings of the National Academy of Sciences}, 117\penalty0
  (29):\penalty0 16880--16890, 2020.

\bibitem[Yu(2013)]{yu2013stability}
Bin Yu.
\newblock Stability.
\newblock \emph{Bernoulli}, 19\penalty0 (4):\penalty0 1484--1500, 2013.

\bibitem[Yu and Kumbier(2020)]{yu2020veridical}
Bin Yu and Karl Kumbier.
\newblock Veridical data science.
\newblock \emph{Proceedings of the National Academy of Sciences}, 117\penalty0
  (8):\penalty0 3920--3929, 2020.

\bibitem[Zhang et~al.(2019)Zhang, Brown, and Cai]{zhang2019semi}
Anru Zhang, Lawrence~D Brown, and T~Tony Cai.
\newblock Semi-supervised inference: General theory and estimation of means.
\newblock \emph{Annals of Statistics}, 47\penalty0 (5):\penalty0 2538--2566,
  2019.

\bibitem[Zhang and Yu(2005)]{zhang2005boosting}
Tong Zhang and Bin Yu.
\newblock Boosting with early stopping: Convergence and consistency.
\newblock \emph{The Annals of Statistics}, 33\penalty0 (4):\penalty0
  1538--1579, 2005.

\bibitem[Zhang and Bradic(2019)]{zhang2019high}
Yuqian Zhang and Jelena Bradic.
\newblock High-dimensional semi-supervised learning: in search for optimal
  inference of the mean.
\newblock \emph{to appear in Biometrika}, 2019.

\bibitem[Zhu and Bradic(2018)]{zhu2018significance}
Yinchu Zhu and Jelena Bradic.
\newblock Significance testing in non-sparse high-dimensional linear models.
\newblock \emph{Electronic Journal of Statistics}, 12\penalty0 (2):\penalty0
  3312--3364, 2018.

\end{thebibliography}

\end{document}